\documentclass{IEEEtran}
\usepackage{amsmath,amsfonts}
\usepackage{algorithmic}
\usepackage{algorithm}
\usepackage{array}
\usepackage{textcomp}
\usepackage{stfloats}
\usepackage{url}
\usepackage{verbatim}
\usepackage{graphicx}
\usepackage{cite}
\usepackage[hidelinks]{hyperref}
\usepackage{booktabs}

\begin{document}

\title{Fast Searching of Extreme Operating Conditions for Relay Protection Setting Calculation Based on Graph Neural Network and Reinforcement Learning}

\author{
    Yan~Li, %
    Jingyu~Wang,~\IEEEmembership{Member,~IEEE}, %
    Jiankang~Zhang, %
    Huaiqiang~Li, %
    Longfei~Ren, %
    Yinhong~Li,~\IEEEmembership{Senior~Member,~IEEE}, %
    Dongyuan~Shi,~\IEEEmembership{Senior~Member,~IEEE}, %
    Xianzhong~Duan,~\IEEEmembership{Senior~Member,~IEEE}%
    \thanks{
      The work was supported by the National Natural Science Foundation of China (Grant No. 52207107). %
      \emph{(Corresponding author: Jingyu Wang)}
    }%
    \thanks{
      Yan Li, Jingyu Wang, Yinhong Li, and Dongyuan Shi are with the State Key Laboratory of Advanced Electromagnetic Technology and the School of Electrical and Electronic Engineering, Huazhong University of Science and Technology, Wuhan, China (e-mails: \href{xmlee@hust.edu.cn}{xmlee@hust.edu.cn}, \href{jywang@hust.edu.cn}{jywang@hust.edu.cn}, \href{liyinhong@hust.edu.cn}{liyinhong@hust.edu.cn}, \href{dongyuanshi@hust.edu.cn}{dongyuanshi@hust.edu.cn}). 
      
      Jiankang Zhang, Huaiqiang Li, and Longfei Ren are with the Northwest Branch of State Grid Corporation of China, Xian, 710048, Shaanxi, China (e-mails: \href{zhangjk@nw.sgcc.com.cn}{zhangjk@nw.sgcc.com.cn}, \href{lihq@nw.sgcc.com.cn}{lihq@nw.sgcc.com.cn}, \href{renlongfei@qq.com}{renlongfei@qq.com}).
      
      Xianzhong Duan is with the College of Electrical and Information Engineering, Hunan University, Changsha, China (e-mail: \href{duanxz@hnu.edu.cn}{duanxz@hnu.edu.cn}).
    }%
}
\maketitle

\begin{abstract}
Searching for the Extreme Operating Conditions (EOCs) is one of the core problems of power system relay protection setting calculation. The current methods based on brute-force search, heuristic algorithms, and mathematical programming can hardly meet the requirements of today's power systems in terms of computation speed due to the drastic changes in operating conditions induced by renewables and power electronics. This paper proposes an EOC fast search method, named Graph Dueling Double Deep Q Network (Graph D3QN), which combines graph neural network and deep reinforcement learning to address this challenge. First, the EOC search problem is modeled as a Markov decision process, where the information of the underlying power system is extracted using graph neural networks, so that the EOC of the system can be found via deep reinforcement learning. Then, a two-stage Guided Learning and Free Exploration (GLFE) training framework is constructed to accelerate the convergence speed of reinforcement learning. Finally, the proposed Graph D3QN method is validated through case studies of searching maximum fault current for relay protection setting calculation on the IEEE 39-bus and 118-bus systems. The experimental results demonstrate that Graph D3QN can reduce the computation time by 10 to 1000 times while guaranteeing the accuracy of the selected EOCs.
\end{abstract}


\section{Introduction}
\label{sec:Introduction}

\IEEEPARstart{R}{elay} protection is one of the most important methods of ensuring the safety and stability of power systems. During the operation of the power system, power equipment can fail at any time. If the faulty equipment is not isolated in time, equipment damage and cascading failures will likely occur. Relay protection serves as a critical technology to trip faulty equipment quickly. In general, the operating condition of power systems changes from time to time, but the settings of protective relays are not easy to change frequently. Therefore, the settings must be determined through a rigorous calculation process to ensure that they can accommodate the ever-changing operating conditions in practical power systems. To ensure that the relay will operate correctly under all system conditions and to avoid the problems of misoperation and refusal to operate, it is necessary to take Extreme Operating Conditions (EOCs) into account in the setting calculation. EOCs include sensitive EOCs and selective EOCs. The former refers to the operating conditions where it is most difficult to meet the sensitivity requirements and the latter refers to those where it is most difficult to meet the selectivity requirements with adjacent protective relays.

Given its significance, the EOC Search (EOCS) problem is arguably one of the most fundamental problems in relay protection setting calculation. It involves searching for the optimum in a high-dimensional discrete space, and hence, belongs to the combinatorial optimization problem \cite{10263956}. Existing methods for solving this problem include brute-force enumeration, heuristic algorithms, and mathematical programming, all of which have a high time complexity related to the dimensionality of the search space \cite{9863874}. With the continuous expansion of power systems, the number of possible operating conditions that need to be evaluated in the EOCS problem grows explosively. Meanwhile, the high penetration of renewable energy and power electronic devices has markedly enhanced the flexibility and randomness of power systems, contributing to more rapid changes in operating conditions. The above situations pose severe challenges on the computation speed of existing methods. It is imperative to investigate a more efficient method to solve the EOCS problem.

Deep Reinforcement Learning (DRL) has flourished in recent years and has been used in various power system applications, such as emergency control \cite{chen_distributed_2024}, optimal power flow \cite{yan_real-time_2020, 9944164}, voltage stability control \cite{kamruzzaman_deep_2021}, and automatic generation control \cite{huang_control_2021}. It is particularly good at finding optimal policies in complex environments and large state spaces, and has therefore been explored for solving combinatorial optimization problems \cite{9523517}. Unlike existing methods that require long online delays, DRL divides the problem solving into offline and online phases. After training intelligent agents in the offline phase, optimal solutions can be found efficiently in the online phase, thereby showing a strong potential to reduce the time overhead of the EOCS problem.

This paper proposes a novel DRL-based method for the fast search of EOCs. Named Graph Dueling Double Deep Q Network (Graph D3QN), the proposed method combines DRL and Graph Neural Networks (GNNs) to offer a significant reduction in the time overhead and comparable accuracy to the previous work. The main contributions of this paper include:
\begin{enumerate}
    \item{\textbf{Applying DRL to the EOCS problem:} To the best of the authors' knowledge, this paper is the first in the literature that applies DRL to solve the EOCS problem. The proposed method alleviates the stress in the computation speed of existing methods and extends the use cases of DRL in current power systems.}
    \item{\textbf{A novel Graph D3QN architecture:} The integration of GNNs into the DRL framework allows to fully utilize the powerful feature extraction capabilities of GNNs to learn the pertinent information of the power system, thereby facilitating more effective policy-making.}
    \item{\textbf{Two-stage training for fast convergence:} A two-stage Guided Learning and Free Exploration (GLFE) training strategy is proposed to use a small amount of enumeration to instruct the inference of the DRL model to achieve high accuracy and stable convergence.}
    \item{\textbf{Validation in relay setting calculation scenarios:} The proposed Graph D3QN method is implemented in relay setting calculation scenarios for case studies. Experimental results show that the proposed method has a 1\%-accuracy of over 98\%, with the computation time reduced by 10 to 1000 times.}
\end{enumerate}

The rest of the paper is organized as follows. Section \ref{sec:RELATED WORK} introduces the related work, including common setting calculation methods, combinatorial optimization and applications of DRL in power systems. Section \ref{sec:PRELIMINARY TECHNOLOGY} introduces preliminary knowledge about short-circuit calculations and DRL. Section \ref{sec:Graph D3QN} elaborates on the proposed Graph D3QN architecture and the GLFE training framework. Section \ref{sec:EXPERIMENTS} validates the performance of Graph D3QN in solving the EOCS problem. Section \ref{sec:CONCLUSION} concludes the paper.

\section{Related Work}
\label{sec:RELATED WORK}
This section first introduces two different modes of relay protection setting calculation. A brief review of the existing methods for solving combinatorial optimization problems and DRL applications in power systems is then presented.

\subsection{Relay Protection Setting Calculation}
\label{sub:setting}
There are currently two modes of relay protection setting calculation. The first is to calculate the preparatory quantities first and then perform the setting calculation using manual rules. Different protection principles require different preparatory quantities, such as maximum/minimum fault current and zero-sequence fault current. The calculation of preparatory quantities typically involves finding the EOCs.
In \cite{1256359}, a method for computing the Zone-2 settings of distance relays is presented. The method first calculates the impedance seen by the distance relay and then performs the setting calculation, which increases the protection range while avoiding causing coordination problems.
In \cite{7359149}, an overcurrent protection method is proposed, which first calculates boundary scaling wavelet coefficient energy via the boundary discrete wavelet transform, and then makes judgments with neutral current and current phase taken into consideration.

The second mode is to take the settings of protective relays of the whole system as optimization variables, to approach the objective such as the shortest total protection operating time. 
In \cite{9211731}, the setting coordination problem is modeled as a mixed-integer linear programming problem, where an overcurrent protection scheme is configured for each setting group to achieve the shortest sum of delays.
In \cite{10373924}, an optimal scheme containing three types of standard relay characteristics is introduced for dual-setting overcurrent relays to make them adapt to both the grid-connected mode and the islanded mode. The optimization problem is solved by the genetic algorithm and grey wolf algorithm.
In \cite{10587055}, a protection setting scheme is proposed to minimize the total running time by configuring the optimum number of relays.

Comparing the two modes above, it can be seen that the first mode typically adopts a level-by-level, segment-by-segment coordination strategy between adjacent protection relays. Although the result is not guaranteed to have globally optimal performance, a set of reasonable settings can be determined in a relatively short time. Although the second mode of global optimization may result in better overall performance, the computational time tends to be unacceptable for large power systems with so many optimization variables and constraints. Therefore, the first mode is generally used in practical engineering.

\subsection{Combinatorial Optimization}
Individual devices in the power system need to be turned on and off in the trial to find the extreme preparatory quantities, which is inherently a combinatorial optimization problem. Besides brute-force enumeration, heuristic algorithms and mathematical programming are the two mainstream methods for solving combinatorial optimization problems.
\subsubsection{Heuristic algorithms}
    Heuristic algorithms are a type of problem-solving algorithm used to find good solutions to complex problems. 
    In \cite{1425605}, a multi-agent particle swarm optimization algorithm is proposed to solve the reactive power dispatch problem. Each agent lives in a grid-like environment, competes and cooperates with its neighbors, and learns knowledge from its neighbors.
    In \cite{6423901}, a learning-based control scheme combining GA with a memory for optimal voltage control is proposed. Past experiences are stored in memory to speed up the search for GA and improve the quality of the solutions.
    In \cite{9870563}, a fast identification algorithm based on greedy policy is designed to identify the vulnerable set and avoid cascading events. The algorithm exploits the strong correlation between different $N-k$ states and is easy to take into account prior knowledge.
    However, heuristic algorithms have the common disadvantage that they are prone to fall into local optimums and converge too early in the search process \cite{NGUYEN2015233}. In addition, heuristic algorithms often do not have a strong mathematical foundation.
    
\subsubsection{Mathematical programming}
    In contrast to heuristic algorithms, mathematical programming can provide optimality guarantees.
    In \cite{8928954, 10019591}, the service restoration problems are formulated as mixed-integer second-order cone programming and mixed-integer linear programming problems, respectively.
    In \cite{8892494}, An approximation algorithm is presented for solving the ac optimal power flow problem, which can find a solution close to the optimum in polynomial time.
    While mathematical programming can produce highly interpretable and accurate solutions, it is too inefficient to apply to large-scale problems.

\subsection{DRL Applications in Power Systems}
    DRL techniques have been applied in a variety of contexts in the field of power systems. In \cite{chen_distributed_2024}, a hierarchical reduction reinforcement learning framework is proposed for emergency control to ensure transient stability of power systems. The framework includes a self-supervised training mechanism to simultaneously train a critical action identification network and an action strategy determination network. A distributed experience sharing architecture is also integrated into the framework.
    In \cite{yan_real-time_2020} and \cite{9944164}, a Lagrangian-based DRL and a convex-constrained soft actor-critic DRL are proposed to solve real-time optimal power flow problems. 
    In \cite{kamruzzaman_deep_2021}, a multi-intelligent deep reinforcement learning framework is proposed to increase the resilience of the power system through shunt reactive power compensators. DRL is used to determine the optimal location and compensation capacity of devices in the event of multi-line faults.
    In \cite{huang_control_2021}, an energy storage system control model using DRL in the scene of the combined wind-solar storage system is proposed to realize the coordination operation of wind power and photovoltaic power.

\section{Preliminary Knowledge}
\label{sec:PRELIMINARY TECHNOLOGY}
This section presents the fundamental principles of DRL, including an overview of the components of Reinforcement Learning (RL), the Deep Q Network (DQN), and the GNN.

\subsection{Reinforcement Learning}
RL addresses the issue of optimizing cumulative reward by enabling agents to learn values and strategies through interaction with their environment\cite{1996Reinforcement}. As shown in Fig. \ref{fig:RL_F}, a typical RL framework comprises the following components:

\begin{figure}[!t]
\centering
\includegraphics[width=3.5in]{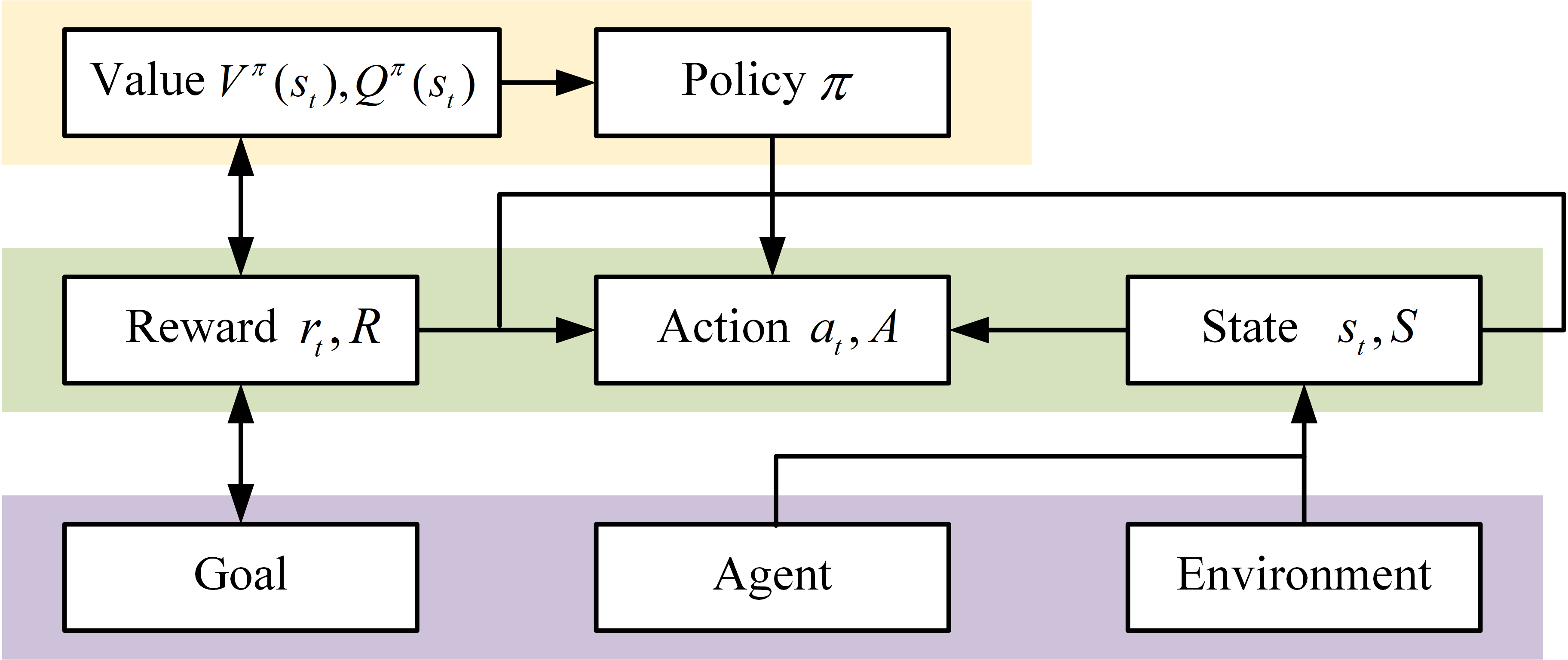}
\caption{The components of an RL framework.}
\label{fig:RL_F}
\end{figure}

\begin{enumerate}
\item{Environment, agent and goal. The agent optimizes parameters by continuously interacting with the environment and gradually reaches the goal.}
\item{$s_t$, $a_t$, $r_t$ represent the state of the environment, the action chosen by the agent, and the reward obtained at the moment $t$. $S$, $A$, $R$ represent the state space, the action space and the reward function, respectively.}
\item{The policy function is $\pi:S\rightarrow A$. The optimal policy is denoted as $\pi^\ast$.}
\item{The state value function $V^\pi(s_t)$ and the state action value function $Q^\pi\left(s_t,a_t\right)$. Value is defined as the expectation of the sum of future rewards. The goal of the agent is to find the optimal policy maximizing the value.}
\end{enumerate}

\subsection{D3QN}
\label{sub:D3QN}
D3QN is a value-based deep reinforcement algorithm, which evaluates the value function and selects actions with the maximum value to execute. The most basic value-based algorithms are Q-Learning \cite{Christopher1992Q} and DQN \cite{2013Playing}, which estimate the value of actions by value tables and value networks respectively. The iterative formula is shown as
\begin{equation}
{Q\left(s,a\right)\gets (1-\alpha )Q\left(s,a\right)+\alpha\left[r+\gamma\max_{a\in A}{Q\left(s^\prime,a\right)}\right]} 
\label{eq:DQN}
\end{equation}
where $\gamma$ and $\alpha$ are the decay rate and learning rate. D3QN has two major improvements over the classical DQN, including the integration of the Dueling DQN\cite{2015Dueling} and the Double DQN\cite{2015Deep}. Dueling DQN allows independent learning of the state value and the action value to improve the stability of the RL process, and the iterative formula is shown as
\begin{equation}
Q\left(s,a\right)=V\left(s\right)+A\left(s,a\right)-\frac{1}{\left|A\right|}\sum_{a} A\left(s,a\right)
\label{eq:duelingDQN}
\end{equation}

Double DQN uses two neural networks with identical structures but different parameters to calculate the target value and predicted value of $Q$ to address the overestimation in DQN and can be formulated as
\begin{equation}
\begin{aligned}
Q^{p}\left(s,a\right) \gets & (1-\alpha)Q^{p}\left(s,a\right) + \\ 
& \alpha\left[r + \gamma Q^{t}\left(s^\prime,  \underset{a\in A}{\mathrm{argmax}}{Q^{p}\left(s^\prime,a\right)}\right)\right]
\end{aligned}
\label{eq:DDQN}
\end{equation}
where $Q^{p}$ is the prediction network and $Q^{t}$ is the target network. Therefore, the update formula for D3QN is
\begin{equation}
\begin{aligned}
\left\{\begin{array}{l}
Q^{p}\left(s,a\right) \gets (1-\alpha)Q^{p}\left(s,a\right) + \\
 ~~~~~~~~~~~~~~~\alpha\left[ r + \gamma Q^{t}\left(s^\prime, \underset{a\in A}{\mathrm{argmax}}{Q^{p}\left(s^\prime,a\right)}\right)\right]
 \\
Q^{p}\left(s,a\right) = V^{p}\left(s\right)+A^{p}\left(s,a\right)-
\frac{1}{\left|A^{p}\right|} \underset{a}{\sum} A^{p}\left(s,a\right)
 \\
Q^{t}\left(s,a\right) = V^{t}\left(s\right)+A^{t}\left(s,a\right)-
\frac{1}{\left|A^{t}\right|} \underset{a}{\sum} A^{t}\left(s,a\right)
\end{array}\right.
\end{aligned}
\label{eq:D3QN}
\end{equation}

\subsection{Graph Neural Network}
\label{sub:GNN}
GNN\cite{2009TheGNN} is widely used for processing graph-structured data. The principle of GNN is to update the features of a node by aggregating the features of all connected nodes, so that the information is propagated in the topological space.
Multiple GNN layers are usually stacked to extract complex features.

Graph Convolutional Network (GCN) is a basic GNN that aggregates the neighboring nodes information through graph convolution operations \cite{Kipf2016SemiSupervisedCW}. The updated features of nodes can be computed as
\begin{equation}
\begin{aligned}
\mathbf{H}^{(l+1)} & = \sigma\left( \hat{\mathbf{A}} \mathbf{H}^{(l)} \mathbf{W}^{(l)} \right) \\
& = \sigma\left( \tilde{\mathbf{D}}^{(-\frac{1}{2})} \tilde{\mathbf{A}} \tilde{\mathbf{D}}^{(-\frac{1}{2})} \mathbf{H}^{(l)}\mathbf{W}^{(l)}\right)
\end{aligned}
\label{eq:GNN}
\end{equation}
where $\mathbf{H}^{(l)}$ represents the node feature matrix of the $l$-th layer; $\tilde{\mathbf{A}}$ is the adjacency matrix with self-loops; $\tilde{\mathbf{D}}$ is the node degree matrix; $\mathbf{W}$ is the weight matrix; and $\hat{\mathbf{A}}=\tilde{\mathbf{D}}^{(-\frac{1}{2})} \tilde{\mathbf{A}} \tilde{\mathbf{D}}^{(-\frac{1}{2})}$ represents the symmetrically normalized adjacency matrix. $\sigma(\cdot)$ denotes the activation function.

\section{Proposed Graph D3QN Framework}
\label{sec:Graph D3QN}
D3QN, as a mature algorithm in value-based RL, can effectively handle problems with discrete state and action spaces\cite{9904958}. In addition, power systems can be represented as graph-structured data comprising nodes and edges, whose indices are unordered and can be designated arbitrarily. This feature makes learning models that can only learn in the Euclidean space less suitable for dealing with power system problems. In contrast, GNN treats every node and every edge in the power system evenly and can directly learn knowledge in the topological space. Therefore, this paper considers combining GNN and D3QN as Graph D3QN, where the GNN is used to extracts information from the power system, and the DRL agent selects EOCs according to the information. This section elaborates the Graph D3QN framework for the EOCS in relay protection setting calculation.

\subsection{Mathematical Model of EOCS}
\label{sub:EOCS math model}
\begin{figure}[!t]
\centering
\includegraphics[width=3.5in]{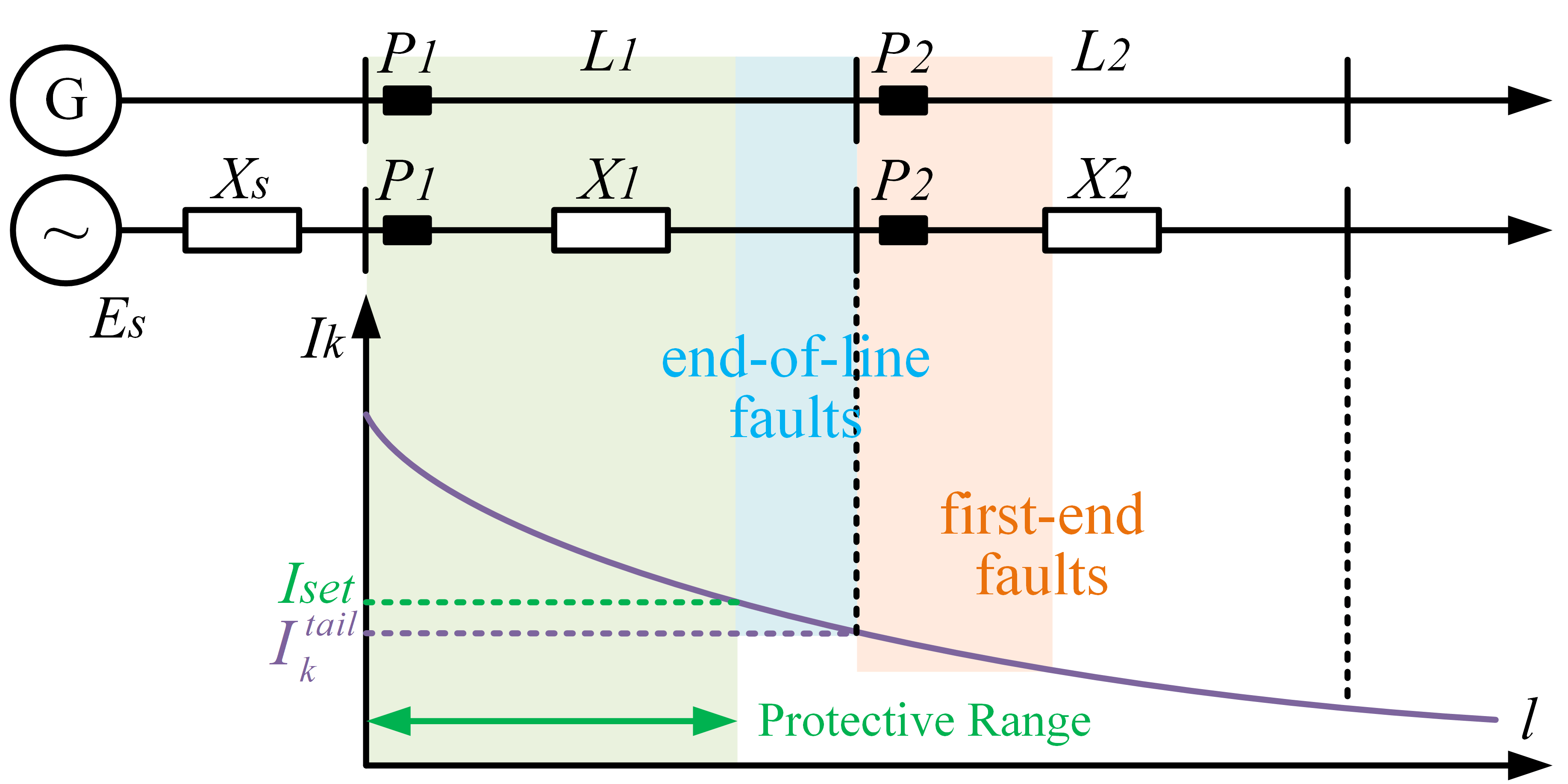}
\caption{Principle of Overcurrent Protection Setting.}
\label{fig:overcurrent}
\end{figure}

In this paper, the first mode of setting calculation described in Section \ref{sub:setting} is adopted, i.e., the setting preparatory quantities are derived before setting calculation. There are many types of protection applied in the power system, including overcurrent protection\cite{5871328}, distance protection\cite{7339488}, zero-sequence protection\cite{6860231}, and so on. Instantaneous overcurrent protection is a simple, reliable, and widely used protection for defending against short-circuit faults on transmission lines. Taking the power system in Fig. \ref{fig:overcurrent} as an example, when a short-circuit occurs at $l$, the short-circuit current $I_k$ flowing through the line is
\begin{equation}
    I_k=\frac{E_s}{X_s+X_k}= \frac{E_s}{X_s+X_ul}
    \label{eq:maxIk}
\end{equation}
where $E_s$ and $X_s$ are the electromotive force and reactance of the power source, respectively, and $X_u$ is the positive sequence reactance per unit length of the line. To fulfill the protection selectivity, $P_1$ can only be disconnected when short-circuit faults occur on $L_1$. However, measurements with current transformers have errors. To avoid confusion between the fault occurring at the tail-end of the protected line and that occurring at the head-end of the downstream line, the pickup current is set to be slightly higher than the maximum short-circuit current at the end of the line. The setting value of the pickup current $I_{set}$ and the action criterion are as follows
\begin{equation}
    I_{set}=KI_k^{tail,max}
    \label{eq:i_set}
\end{equation}
\begin{equation}
    I>I_{set}
    \label{eq:i_pick_up}
\end{equation}
where $I_k^{tail,max}$ is the maximum fault current occurs at the tail-end of the protected line and $K>1$ is a manually set coefficient, usually taken as 1.2 to 1.3. Therefore the protective range of the overcurrent protection $P$ is the part in green in Fig. \ref{fig:overcurrent}. The selective EOCS problem in this scenario is hence to find the operating state of other power equipment so that the tail-end fault current reaches its global maximum. Therefore, $I_k^{tail,max}$ is the preparatory quantities needed to be taken before setting the instantaneous overcurrent protection. For the sake of brevity, this paper only considers searching in the combination of on/off states of other transmission lines in the power system. 

It is empirically established that the farther an equipment state change is from the fault point, the less effect it has on the fault current. This inverse distance principle is employed in traditional relay protection setting calculation methods for search EOCs. 
Local enumeration is the most representative and commonly used method\cite{7496931}. The method searches in the vicinity of the fault at level $r$, called the searching region $R$, and considers at most $N-k$ cases, i.e., involving at most $k$ devices out of service within $R$. Then applies brute-force enumeration to the combinations of the on/off states of devices within $R$, to find the operating conditions that make the fault current maximum as the EOC.
The downside of this method is that the designation of the searching region $R$ is mostly based on empirical evidence. A too small region is insufficient to identify global EOCs, whereas a too large region will cause an increased time overhead.

To address the above disadvantage, this section introduces a fast EOCS method for calculating the setting of instantaneous overcurrent protection based on Graph D3QN. In practice, three-phase short-circuit faults generally lead to the most severe consequences, while the short-circuit current is negatively correlated with the ground impedance. The following design aims to identify the EOC that maximizes the fault current in the case of a three-phase and zero grounding impedance short-circuit fault at the tail-end of a line. A mathematical model of the EOCS for instantaneous overcurrent protection can be established as
\begin{equation}
    I_k^{tail,max}=\max_{\tau \in T} I_{k,\tau,z,f}^{tail} =\max_{\tau \in T} SCC(x_1,...,x_m,z,f)
    \label{eq:maxIk}
\end{equation}
\begin{equation}
s.t.\left\{\begin{array}{l}
\tau=\left \{ x_1,...,x_m \right \} 
 \\
x_i \in \left \{ 0,1 \right \} ,i \in \left \{ 1,2,...,m \right \} 
\\
z = 0  ,f =f^{(3)}
\end{array}\right.
\label{eq:yueshu}
\end{equation}
where $T$ is the topological space of the power system and $\tau$ is one of the operating conditions in $T$. $x_1,...,x_m$ represent the start/stop states of the transmission lines, $z$ is the ground impedance at the fault point, $f$ is the fault type, and $SCC$ is the function of calculating the short-circuit current flowing through the faulted line.

\subsection{Architecture of Graph D3QN}
\label{sub:Network}

\begin{figure*}[!t] 
\centering 
\includegraphics[width=0.95\textwidth]{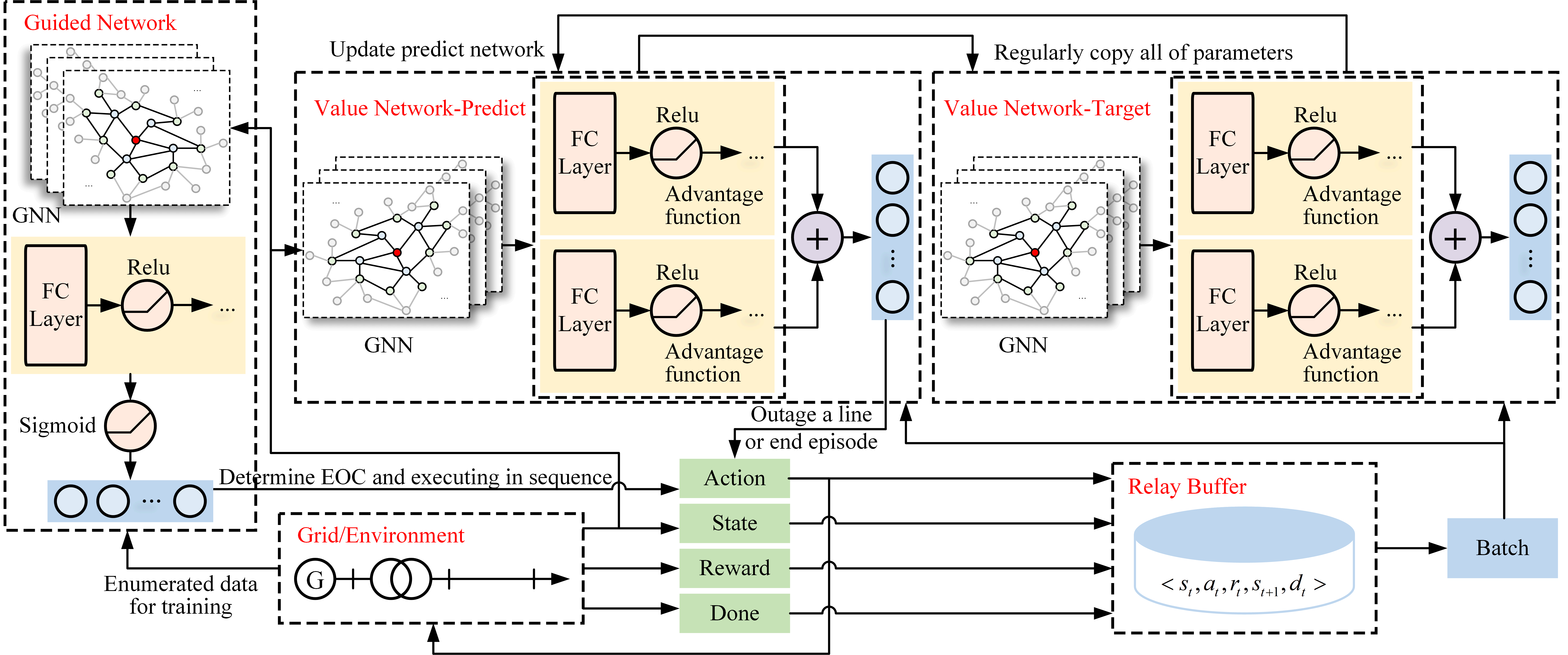} 
\caption{The overall architecture of Graph D3QN.} 
\label{fig:OverallModel}
\end{figure*}

Solving a problem with DRL requires representing the problem as a Markov decision process. Therefore, it is necessary to partition the EOCS problem into a stepwise execution process. First, the initial operating state of the system, i.e., the initial on/off status of all lines, is randomly selected and set as the initial state of the environment. The agent chooses to trip a line based on the protection location and receives a reward. The state of the environment is then updated according to the agent's action. Ideally, this process should be repeated until the agent chooses an action that leaves the environment state unchanged, which means that the system operating condition arrives at the EOC. In practice, however, it is a convention that at most $k$ lines are considered to be triggered simultaneously, where $k$ is a pre-specified parameter. This is reasonable because the probability that faults occur simultaneously on too many lines is very low. 

The definitions of state, action, and reward in this scenario are as follows:
\begin{enumerate}
    \item{\textbf{State:} the system configuration and the current operating condition, including the location of protection, self-loop adjacency matrix, node impedance matrix, and minimum electrical distance matrix. The elements in the minimum electrical distance matrix represent the minimum electrical distance between two buses and are calculated by the Dijkstra algorithm.}
    \item{\textbf{Action:} the action space of the agent is to select a non-protected in-service line for tripping, or to keep the current operating condition unchanged.}
    \item{\textbf{Reward:} the difference between the fault current before and after the action of the agent. If the fault current increases, the difference is positive and the reward is positive. Conversely, if the fault current decreases, the reward is negative and becomes a penalty.}
\end{enumerate}

Graph D3QN comprises two networks, a guide network and a value network. Both are composed of graph convolutional layers and fully connected layers, as shown in Fig. \ref{fig:OverallModel}.
The two networks have the same input, i.e., the state of the environment. For a system comprising $n$ nodes, the length of the feature vector for each node is $3n$, where the first $n$ dimensions signify the topological connection, the second $n$ dimensions represent self-impedance and mutual-impedance, and the last $n$ dimensions represent the minimum electrical distance between a bus and other buses.

The output dimension of the guide network is the same as the number of lines in the system, with each output element corresponding to a specific line.
Lines whose outputs are less than or equal to 0.5 are not activated and are excluded from being tripped. Otherwise, the $k$ lines with the highest outputs are selected to be tripped to form a predicted EOC. If the number of outputs greater than 0.5 is less than $k$, the predicted EOC is to trip all lines whose outputs are greater than 0.5.

The value network employs the D3QN structure, whereby the state value $V$ and action advantage $A$ are calculated first, and then superimposed to get the state-action value $Q$. For a system containing $m$ transmission lines, the output dimension of the value network is also $m$, with each output element representing the value of tripping the corresponding line under the input state. The line with the maximum value is selected as the action to execute. If the value network selects the line where the protection is located or a line that is out of service, this episode is stopped and the on/off states of all lines at that moment is saved as the EOC. 

\subsection{GLFE Training Framework}
\label{sub:Training Framework}

\begin{figure*}[!t] 
\centering 
\includegraphics[width=0.95\textwidth]{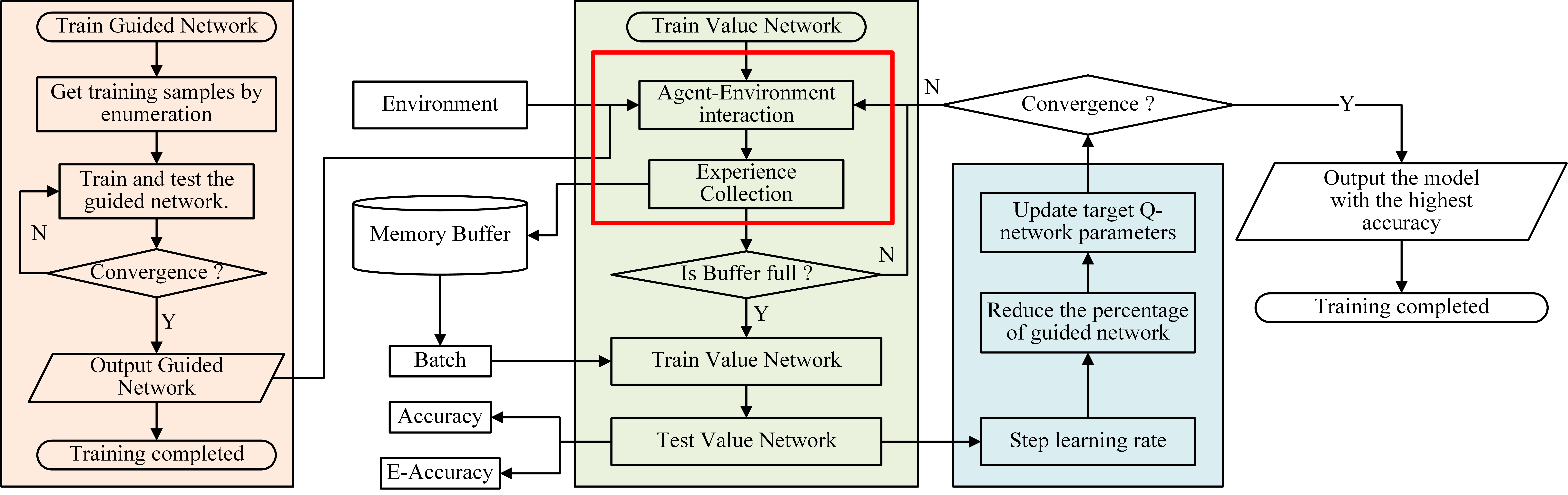} 
\caption{The flowchart of the proposed GLFE training framework.} 
\label{fig:Training_Flowchart}
\end{figure*}

To accelerate the convergence speed, a two-stage GLFE training framework is constructed. First, a small number of sample pairs are generated based on enumeration. The feature of the training samples is the initial state of the system. The label of the samples is the corresponding EOC that leads to the maximum fault current, which is obtained by enumerating the combination of line on/off states. Each element corresponds to the service status of a line, with 1 representing the out-of-service state and 0 for the in-service state. The guide network is trained in a supervised way, using the binary cross entropy as the loss function, to gain a certain degree of EOC prediction ability. Subsequently, the guide network is used to instruct the training of the value network, which aims to facilitate rapid convergence.

The value network is trained with the assistance of the guide network in a semi-supervised manner. During training, a part of label action is provided by the guide network, where $k$ steps of actions are executed sequentially to form a predicted EOC. The other part of label actions is determined by the value network following some exploration strategies. It should be noted that the guide network initially has a relatively high proportion during training, with a subsequent gradual reduction as the training progresses. The loss is calculated by the mean squared error function according to \eqref{eq:D3QN}.

Furthermore, the training process employs the experience replay strategy, whereby all experiences are stored in a buffer in the form of quintuples $<s_t,a_t,r_t,s_{t+1},d_t>$. Once the buffer has reached its capacity, batches are sampled from it for training.
It is crucial to emphasize that the Elements in the EOC are equivalent to each other, independent of the order of actions. Thus, the decay rate $\gamma$ in \eqref{eq:D3QN} is set to 1 to ensure that the sequence of action selection does not influence its value.

\begin{algorithm}[t]
\caption{Extended Exploration Strategy}\label{alg:Exploration Strategy}
\begin{algorithmic}
\STATE {Set the Number $n$ of Actions to Explore.}
\STATE {\textbf{Initialization:} the state \textbf{\textit{s}} of the environment.}
\STATE {Get values of actions by \textbf{Value Network}.}
\STATE {Sort the values and select the top $n$ actions.}
\STATE {\textbf{For} $a$ in actions:}
\STATE \hspace{0.5cm}\textbf{do} $a$
\STATE \hspace{0.5cm}Environment updated to $s'$.
\STATE \hspace{0.5cm}Return reward $r$ and done $d$.
\STATE \hspace{0.5cm}Record $(s, a, s', r, d)$ in Memory Buffer.
\STATE \hspace{0.5cm}\textbf{If} not $d$:
\STATE \hspace{1cm}$n=n-1$
\STATE \hspace{1cm}\textbf{If} $n>0$:
\STATE \hspace{1.5cm}\textbf{Execute} this Strategy under $s'$.
\STATE \hspace{1cm}\textbf{End If}
\STATE \hspace{0.5cm} \textbf{End If}
\STATE \textbf{End for}
\end{algorithmic}
\end{algorithm}

To accelerate training and convergence, an extended exploration strategy is also proposed. The previous action selection strategies were limited to attempting one action under a state, which prolongs training time. The extended exploration strategy employs Algorithm \ref{alg:Exploration Strategy} to attempt multiple actions under a single input state. This strategy is suitable for use in problems where an episode stops after performing only a few rounds of actions. As the episode lengthens, the exploration time will increase significantly. Additionally, the number of actions to be explored, $n$, is a crucial hyperparameter. In this paper, approximately 10\% of the dimension of the action space is explored.

During the training process of the value network, some parameters need to be adjusted in real-time, including the percentage of the guide network selecting actions, the learning rate, and whether the parameters of the target Q network require updating. 
In this paper, it is uniformly set to execute the above operations after the conclusion of one round of training. 
Therefore the training framework of Graph D3QN can be constructed as illustrated in Fig. \ref{fig:Training_Flowchart}.
The GLFE training framework is the most important part of the framework, which is highlighted in red, as presented in Algorithm \ref{alg:GLFE}.

\begin{algorithm}[t]
\caption{Guided Learning and Free Exploration}\label{alg:GLFE}
\begin{algorithmic}
\STATE {Assign Ratios to Guided Learning and Free Exploration.}
\STATE {\textbf{Initialization:} randomize the state $s$ of the environment.}
\STATE {\textbf{If }Guided Learning:}
\STATE \hspace{0.5cm}Get actions by \textbf{Guide Network}.
\STATE \hspace{0.5cm}\textbf{For} action $a$ in actions:
\STATE \hspace{1cm}\textbf{do} $a$
\STATE \hspace{1cm}Environment updated to $s'$.
\STATE \hspace{1cm}Return reward $r$ and done $d$.
\STATE \hspace{1cm} $d=\text{TRUE}$ if $a$ is the final else FALSE.
\STATE \hspace{1cm}Record $(s, a, s', r, d)$ in the buffer.
\STATE \hspace{0.5cm}\textbf{End For}
\STATE {\textbf{Else }Free Exploration:}
\STATE \hspace{0.5cm}\textbf{While $d$:}
\STATE \hspace{1cm}\textbf{Execute} Extended Exploration Strategy under $s$.
\STATE \hspace{0.5cm}\textbf{End While}
\STATE \textbf{End If}
\end{algorithmic}
\end{algorithm}

It is noteworthy that the guide network outputs the predicted EOC directly based on the input system operating, while the value network picks one line out of service at a time, based on the current state of the environment, and selects another line based on the updated state of the environment until the end of the episode state. This implies that the value network employs a recursive process to search for the EOCs.

\section{Experiments and Discussions}
\label{sec:EXPERIMENTS}
In this section, the proposed Graph D3QN is validated in a practical selective EOCS problem used for calculating the setting of instantaneous overcurrent relays. The effectiveness experiments, comparison experiments, and ablation experiments are conducted on the IEEE 39-bus system, while the scalability experiment is implemented on the IEEE 118-bus system.

\subsection{Experimental Setup}
\label{sub:Setup}

\subsubsection{Experiment environment} Experiments are performed on a high-performance server with two Intel Xeon Gold 6248R 3.00 GHz CPUs, 128 GB RAM, and a NVIDIA RTX 3090 GPU. The operating system is Ubuntu Linux 22.04. The proposed Graph D3QN framework is implemented based on Pytorch 2.1.1 with CUDA Toolkit 12.1. The Python interpreter version is 3.10.

\subsubsection{Test scenarios} 
\label{subsub:scenarios}
In practical relay protection setting calculation applications, it is required to search for the EOCs of all relays under a certain initial line on/off status of the system. Two test scenarios are developed to test the accuracy of the proposed method. Scenario 1 is used during the training process, while Scenario 2 is employed at the end of the training, as a supplement to Scenario 1.

Scenario 1: Randomly select $N_1$ cases of input features, including the initial on/off status of lines and the protection location, and compare the extreme fault currents calculated by the value network with those calculated by the existing methods.

Scenario 2: Randomly select $N_2$ cases of the initial on/off status of lines, calculate the extreme fault currents of all protective relays in the system, and compare the results with the existing methods.

\subsubsection{Performance metrics}
For testing the guide network, thousands of system initial operating conditions are taken and input to the guide network. Compare whether the EOC output by the guide network completely overlaps with the one obtained by enumeration. If it completely overlaps, the prediction of the guide network is regarded as correct. Otherwise, it is regarded as wrong. The percentage of correct samples represents the accuracy of the guide network.

For testing the value network, a number of initial operating conditions of the system are taken, with the maximum fault current calculated by the value network compared with the extreme fault current calculated by the enumeration method. If they are equal, the EOC selected by the value network is correct, and the percentage of correctness indicates the accuracy of the value network. In practical applications of relay protection setting calculations, the maximum short-circuit current search is usually allowed to have some error due to the empirical coefficient $K$ in \eqref{eq:i_set}. Therefore, a metric termed $e$-accuracy is also defined, which is used to evaluate the proposed method with an acceptable error $e$ margin taken into account. The $e$-accuracy metric is the percentage of samples whose extreme fault currents produced by the value network exhibit a discrepancy equal to or less than $e$ compared with the extreme fault currents calculated through the enumeration method. In this paper, $e$ is set to 1\%, 2\%, 5\%, respectively.

\subsection{Effectiveness Experiment}
\label{sub:Effectiveness}
Experiments are first conducted on the IEEE 39-bus system to verify the effectiveness of the proposed Graph D3QN in solving the EOCS problem. The system contains 39 buses and 34 transmission lines. 
The initial operating condition of the system is considered to be randomly selected within the range of $N-3$, which means that there are 0 to 3 lines that are not in service. Set the maximum number of tripped lines $k$ to 3. Accordingly, the scope of this experiment will encompass the operating conditions within the range of $N-6$ of the system.

For the effectiveness experiments, the guide network is designed to consist of two graph convolutional layers, two fully connected layers, and finally output by a sigmoid function. The value network consists of two graph convolutional layers and three fully connected layers. The hyperparameters of the Graph D3QN model are determined by the grid search method, shown in Table \ref{tab:Parameters}, and the same model hyperparameters will be used in Section \ref{sub:Comparison} and Section \ref{sub:Ablation}.

\begin{table}[!t]
\caption{Parameters of Graph D3QN\label{tab:Parameters}}
\centering
\setlength{\tabcolsep}{4.2pt}
\begin{tabular}{*{3}{c}||*{3}{c}}
\toprule
Guided Net & 39 & 118 & Value Net & 39 & 118\\
\midrule
Batch & 128 & 256 & Batch & 64 & 64\\
Training set & 12000 & 30000 & $\alpha$, $\varepsilon$ & 0.9, 0.15 & 0.9, 0.2\\
Verify set & 1500 & 4000 & Action num & 3 & 10\\
Test set & 3000 & 8000 & N1, N2 & 1000, 20 & 7000, 40\\
Learning rate & 0.001 & 0.001 & Learning rate & 0.001 & 0.001\\
Train Epochs & 2000 & 2000 & Memory & 10000 & 40000\\
Initial percentage & 0.9 & 0.99 & gamma & $1/\sqrt{10}$ & $1/\sqrt{10}$\\
Percentage step & 0.03 & 0.001 & Step size & 5 & 100\\
\bottomrule
\end{tabular}
\end{table}

\begin{figure}[!t]
\centering
\includegraphics[width=3.5in]{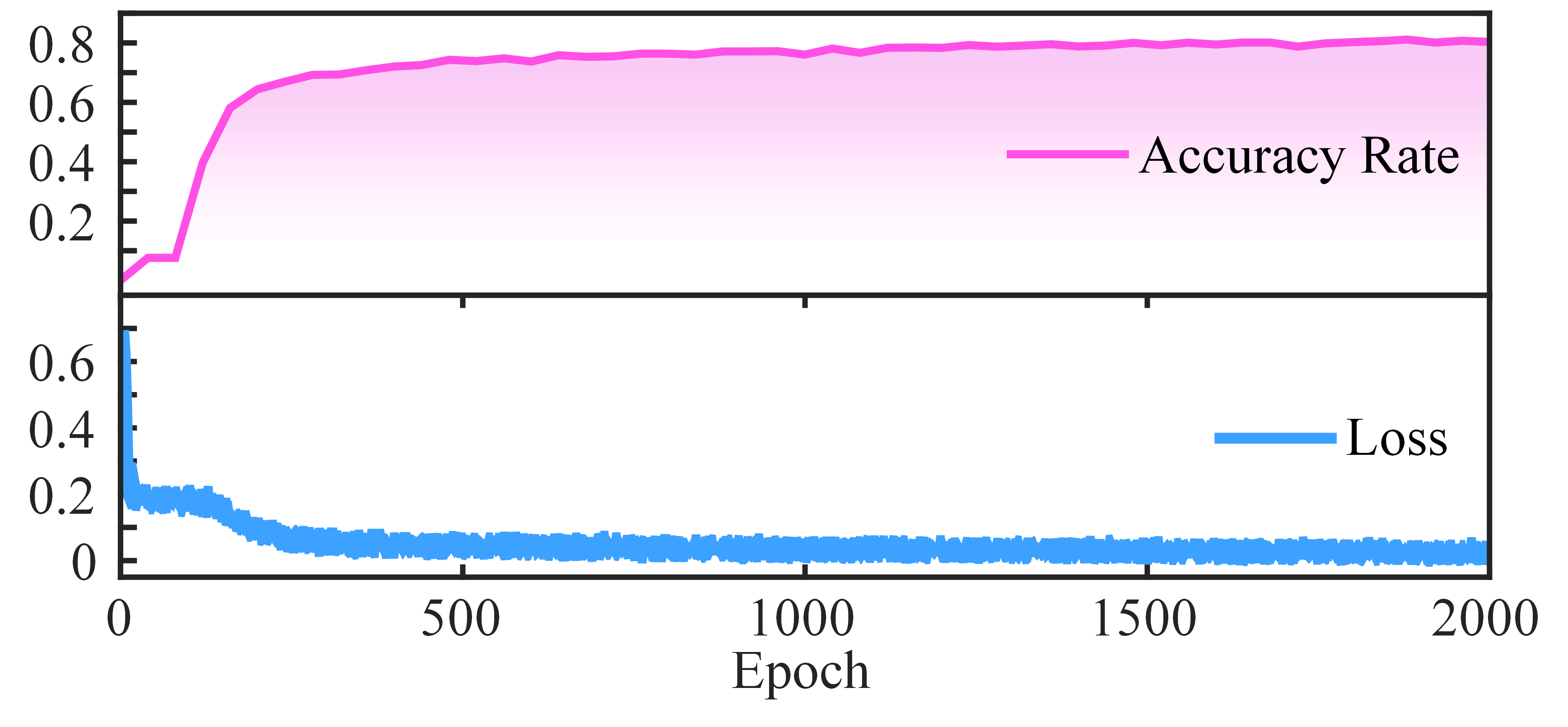}
\caption{Accuracy and loss curves of the guide network on the 39-bus system.}
\label{fig:g_loss_acc_39}
\end{figure}

\begin{figure}[!t]
\centering
\includegraphics[width=3.5in]{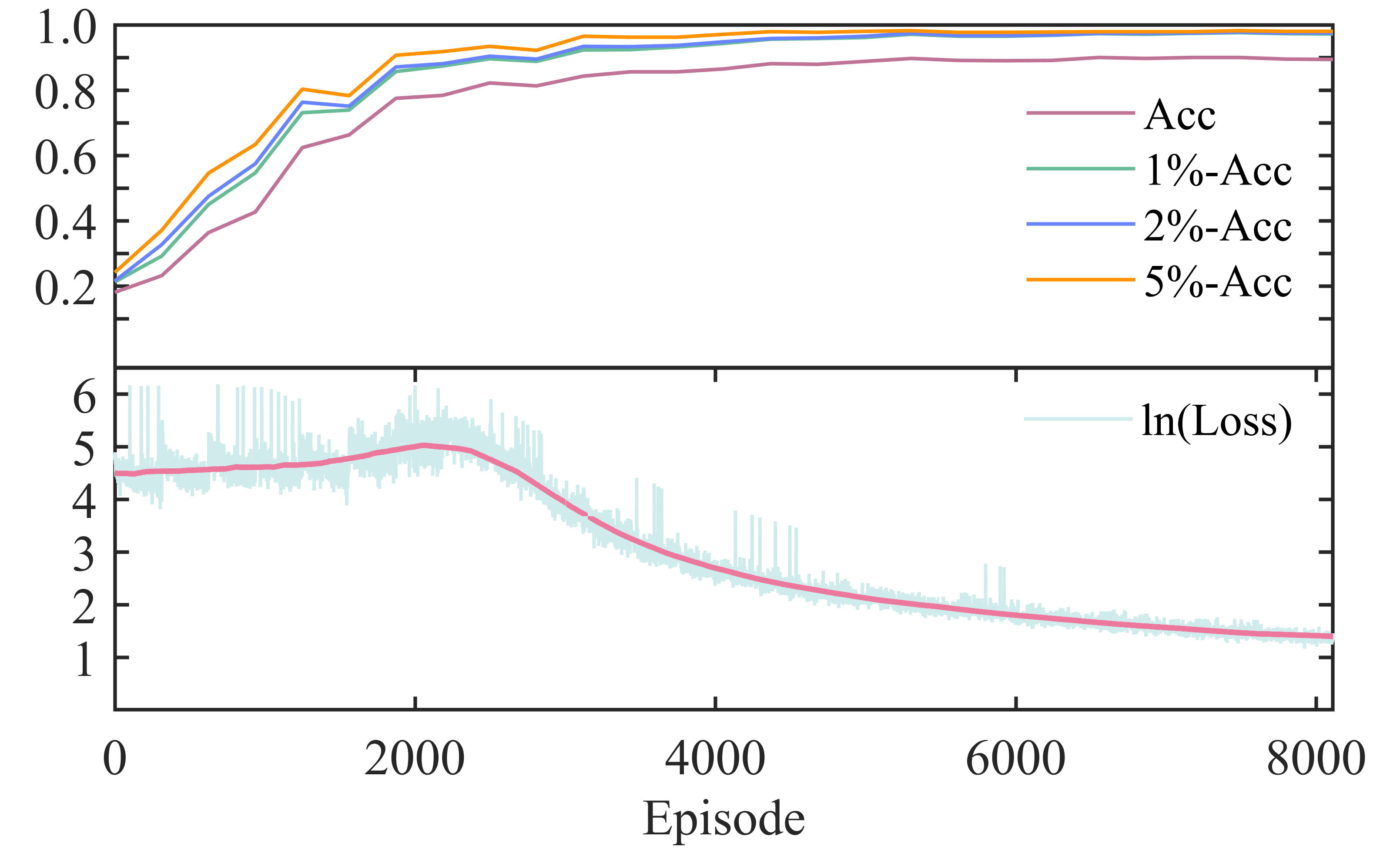}
\caption{Accuracy and loss curves of the value network on the 39-bus system.}
\label{fig:v_loss_acc_39}
\end{figure}

The loss and accuracy curves of the guide network and value network are shown in Fig. \ref{fig:g_loss_acc_39}, respectively. The accuracy of the guide network is 80.3\%, which reflects it has a certain level of EOCS ability. The accuracy of the value network is shown in Table \ref{tab:acc_39} and Fig. \ref{fig:v_loss_acc_39}. In scenario 2, the extreme fault currents were calculated for all protections in 20 initial on/off status of lines, yielding a total of 630 initial system operating conditions. As evidenced by the results, the accuracy of the value network exceeds 90\%, while the $e$-accuracy rates all surpass 98\%. These findings substantiate the efficacy of the proposed DRL-based EOCS-solving method.

\begin{table}[!t]
\caption{Performance of Graph D3QN on IEEE 39-Bus System\label{tab:acc_39}}
\centering
\begin{tabular}{*{5}{c}}
\toprule
Scenario & Accuracy & 1\%-Accuracy & 2\%-Accuracy & 5\%-Accuracy \\
\midrule
1 & 90.20\% & 98.30\% & 98.60\% & 98.80\% \\
2 & 90.635\% & 98.254\% & 99.048\% & 99.524\% \\
\bottomrule
\end{tabular}
\end{table}

Moreover, the total number of possible operating conditions in the $N-6$ range is 47,322,628. However, the semi-supervised GLFE training framework only requires 12,000 samples from brute-force enumeration for training the guide network and 129,601 different states in the $N-6$ range for training the value network. The experienced operating conditions are only 0.273\% of the total $N-6$ cases, but the proposed Graph D3QN finally achieves an accuracy of over 90\%. This also serves as evidence to demonstrate the effectiveness of the proposed EOCS method.

\subsection{Comparison Experiment}
\label{sub:Comparison}
In this section, the proposed Graph D3QN method is compared with the existing EOCS solving methods, including the brute-force enumeration method, local enumeration method, GA, and the combination of the Convolutional Neural Network (CNN) and the D3QN. Scenario 1 is used as the test scenario. The brute-force enumeration method considers all line-tripping cases within the range of $N-3$, where the operating condition with the maximum fault current is identified as the EOC. The local enumeration method only considers tripping lines with a certain number of adjacent levels surrounding the protection installation location. In this experiment, the local neighborhood of 3 levels of adjacency is designated. The GA method models the line on/off status as genetic codes. The fitness function is constructed depending on the fault current, with the EOC finally obtained via the evolution operation. The CNN D3QN framework is similar to the proposed Graph D3QN, except that the GCN structure is replaced with CNN.

\begin{table}[!t]
\caption{Comparison of Different Methods on IEEE 39-Bus System\label{tab:COMPARISON 39}}
\centering
\begin{tabular}{*{5}{c}}
\toprule
Method & Accuracy & 1\%-Accuracy & Computation Time \\
\midrule
Graph D3QN & 90.2\% & 98.3\% & 23.63 ms\\
Global Enum & 100\% & 100\% & 36.48 s\\
Local Enum & 89.3\% & 100\% & 267.88 ms\\
GA & 62.3\% & 81.1\% & 2 min 54 s\\
CNN D3QN & 81.7\% & 91.6\% & 49.58 ms\\
\bottomrule
\end{tabular}
\end{table}

As seen from the results in Table \ref{tab:COMPARISON 39}, the average computation time of the Graph D3QN to process a single sample is 1544 times shorter than that of the global brute-force enumeration method. Meanwhile, the accuracy of Graph D3QN is comparable to that of the local enumeration method, but the computation time is approximately 10 times shorter. The GA has a low accuracy and a long optimization time, which indicates that it is challenging to use heuristic algorithms to solve large-scale EOCS problems. In comparison to the CNN D3QN, the Graph D3QN has been demonstrated to exhibit a higher degree of accuracy. Besides, the training time of Graph D3QN is 1 hour and 33 minutes, which is a little bit shorter than the training time of CNN (1 hour and 54 minutes). This evidence indicates that GNN is more effective than CNN in addressing the EOCS problem.

\subsection{Ablation Experiment}
\label{sub:Ablation}
\begin{figure}[!t]
\centering
\includegraphics[width=3.5in]{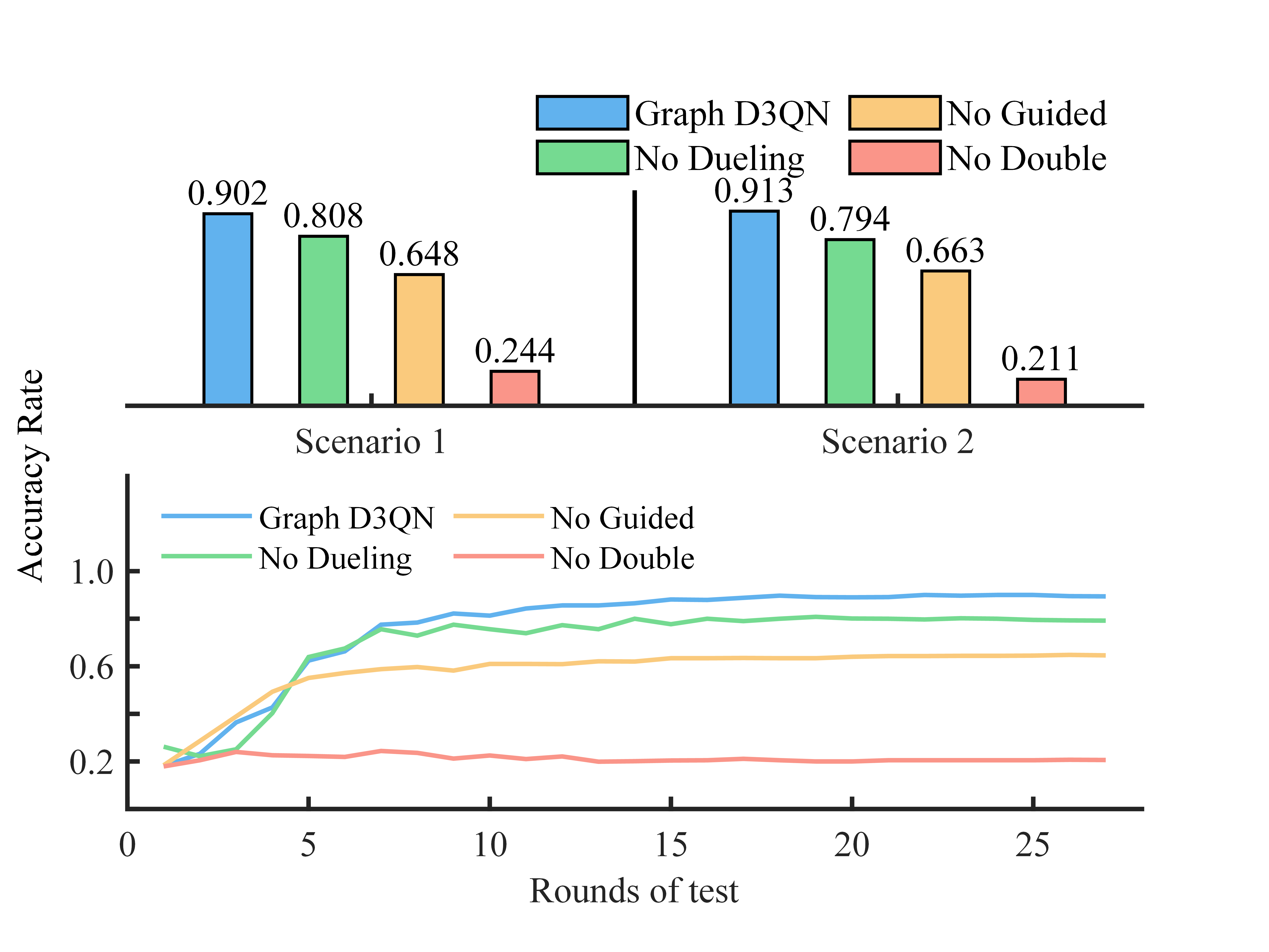}
\caption{Accuracy results of the ablation experiment.}
\label{fig:xiaorong}
\end{figure}

To demonstrate the necessity of each component in the proposed Graph D3QN framework, ablation experiments are performed in this subsection. The guide network, Dueling DQN, and Double DQN in Graph D3QN are removed, respectively. The rest of the parameters are maintained, and the accuracy is tested with both Scenario 1 and 2. The accuracy results of the ablation experiment are shown in Fig. \ref{fig:xiaorong}. 

According to the results, the accuracy curve of removing all three components has dropped, which proves that all the methods have a certain promotion effect on the overall accuracy. From weak to strong, the promotion effect has an order of Dueling DQN, Guide Network, and Double DQN. The most influential improvement to the ordinary DQN is the Double DQN. The removal of the Double DQN tends to cause the framework to disconverge and thus introduces significant accuracy degradation to the proposed method.

\subsection{Scalability Experiment}
\label{sub:118 Experiment}
To validate the scalability of the proposed Graph D3QN method, experiments are conducted on the IEEE 118-bus system. After merging parallel lines, the system contains 118 buses and 169 transmission lines in total. The initial operating condition of the system is considered to be randomly selected within the range of $N-2$. The maximum line tripping limit $k$ is set to 2, which means that EOCS only considers the $N-2$ range starting from the initial state. Therefore, this experiment will at most cover the operating conditions in the range of $N-4$ of the power system.

For scalability experiment, the guide network consists of three graph convolutional layers, three fully connected layers, and finally a sigmoid output layer. The value network consists of four graph convolutional layers and four fully connected layers. The hyperparameters are set as per Table \ref{tab:Parameters}.

\begin{figure}[!t]
\centering
\includegraphics[width=3.4in]{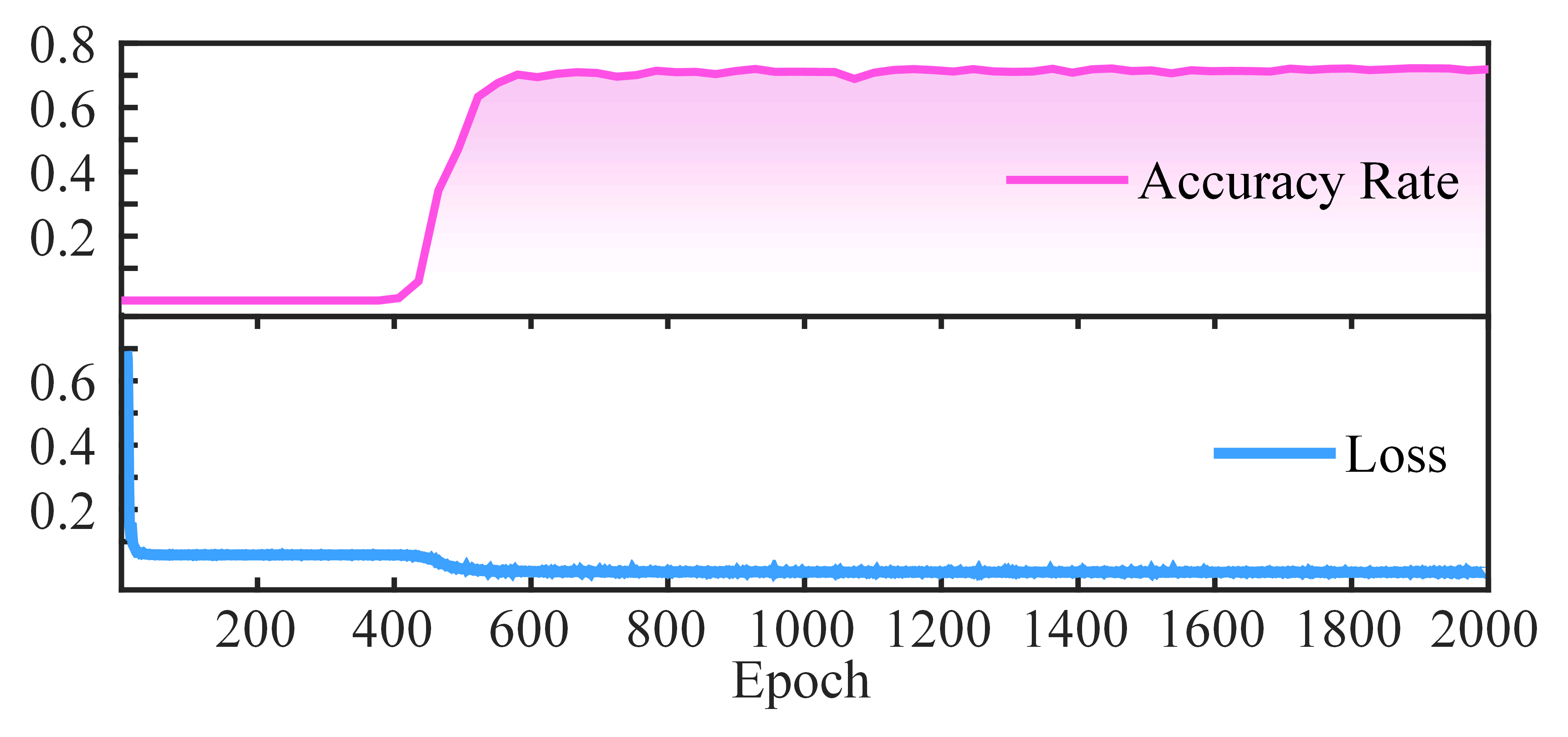}
\caption{Accuracy and loss curves of the guide network on the 118-bus system.}
\label{fig:g_loss_acc_118}
\end{figure}

\begin{figure}[!t]
\centering
\includegraphics[width=3.5in]{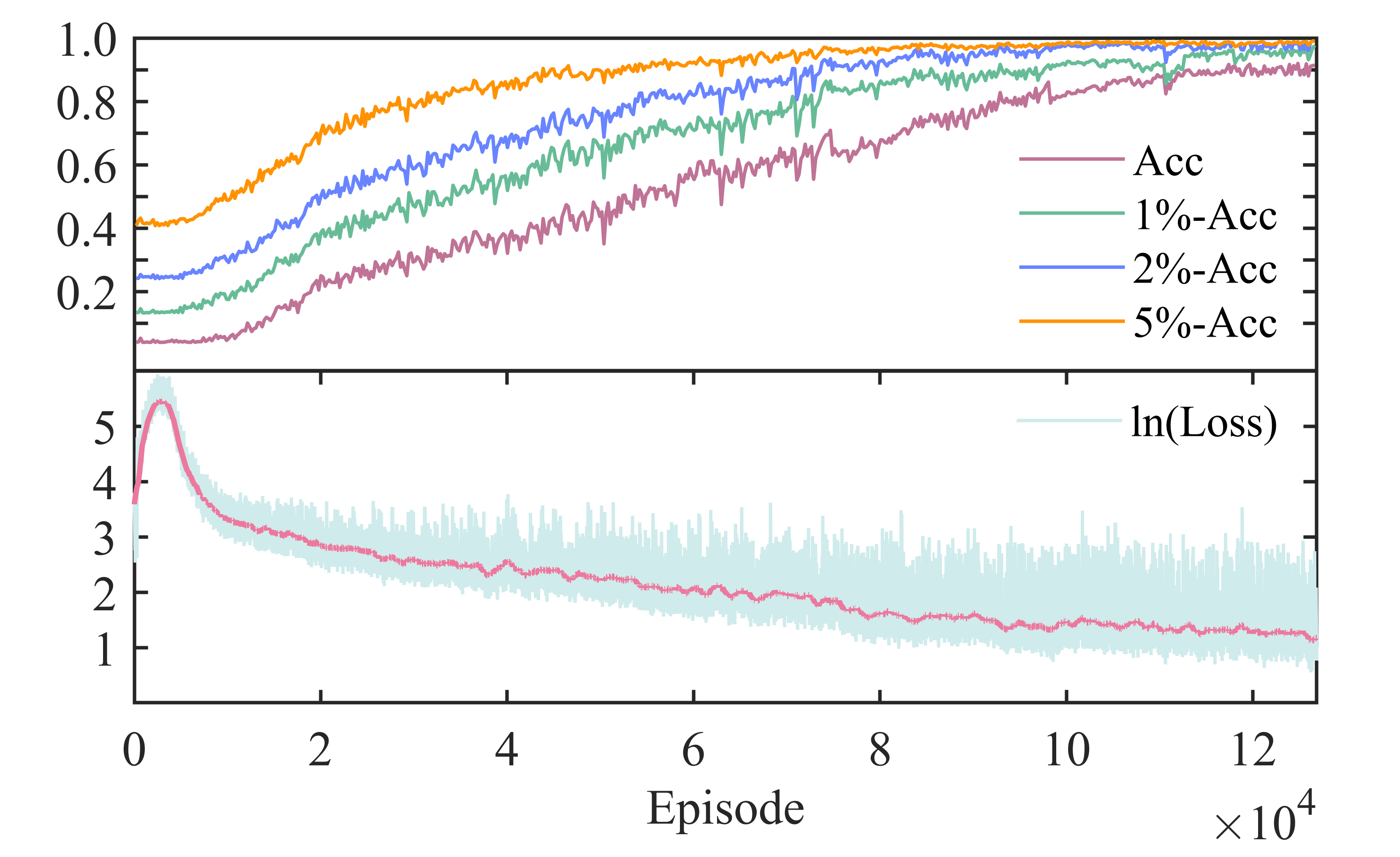}
\caption{Accuracy and loss curves of the value network on the 118-bus system.}
\label{fig:v_loss_acc_118}
\end{figure}

The loss and accuracy curves of the guide network and value network are shown in Fig. \ref{fig:g_loss_acc_118}. The accuracy of the guide network is 72.25\%, which may be considered to indicate some degree of search ability. The accuracy of the value network is shown in Table \ref{tab:acc_118} and Fig. \ref{fig:v_loss_acc_118}. It can be observed that the value network accuracy rate exceeds 91\%, and the $e$-accuracy rates are over 97\%, which proves the effectiveness of the method. Additionally, the total number of states in the $N-4$ range of the 118-bus system is 5,544,778,587, and the value network can achieve over 90\% effectiveness by experiencing only 5,046,487 cases (0.091\%), which also shows the effectiveness and scalability of the proposed method.

As evidenced by the results in Table \ref{tab:COMPARISON 118}, the computation time of the Graph D3QN for a single sample is 1086 times shorter than that of the brute-force enumeration method. The accuracy of the local enumeration method drops seriously while maintaining a high level of $e$-accuracy. This implies that, as the system grows in size, to maintain the same level of accuracy, the local enumeration method must consider a larger range, which in turn increases the computation time.

\begin{table}[!t]
\caption{Performance of Graph D3QN on IEEE 118-Bus System\label{tab:acc_118}}
\centering
\begin{tabular}{*{5}{c}}
\toprule
Scenario & Accuracy & 1\%-Accuracy & 2\%-Accuracy & 5\%-Accuracy \\
\midrule
1 & 91.60\% & 98.20\% & 98.70\% & 99.15\% \\
2 & 91.98\% & 98.70\% & 99.05\% & 99.30\% \\
\bottomrule
\end{tabular}
\end{table}

\begin{table}[!t]
\caption{Comparison of Different Methods on IEEE 118-Bus System\label{tab:COMPARISON 118}}
\centering
\begin{tabular}{*{5}{c}}
\toprule
Method & Accuracy & 1\%-Accuracy & Computing Time\\
\midrule
Graph D3QN & 91.6\% & 98.2\% & 39.24 ms\\
Global Enum & 100\% & 100\% & 42.6 s\\
Local Enum & 29.67\% & 94\% & 1.73 s\\
\bottomrule
\end{tabular}
\end{table}

\subsection{Protection Selectivity Experiment}
\label{sub:Protection Selectivity Experiment}
\begin{figure}[!t]
\centering
\includegraphics[width=3.5in]{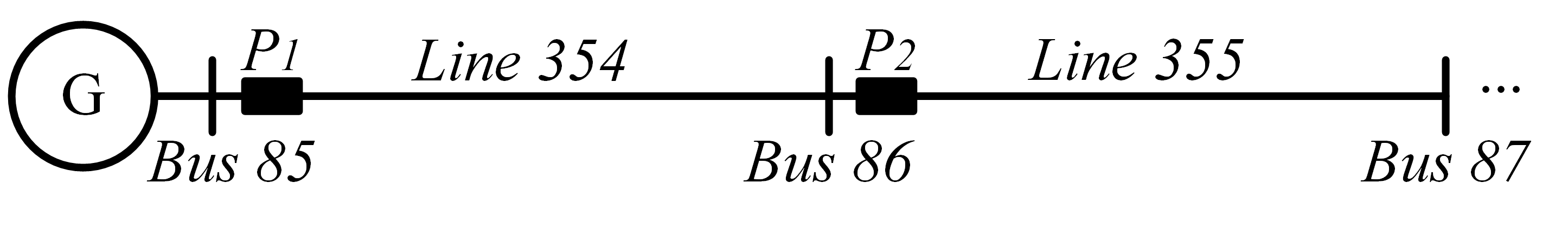}
\caption{Local Topology of IEEE 118-Bus System.}
\label{fig:jubu}
\end{figure}

To verify the practicality of the Graph D3QN, this section will focus on the selectivity of a specific protection. Taking transmission lines 354 and 355 in the IEEE 118-bus system as an example, the local topology of the system is shown in Fig. \ref{fig:jubu}. From the introduction in Section \ref{sub:EOCS math model}, to satisfy the selectivity, it is necessary to make the setting value $I_{set}$ of the protection $P_1$ at the first end of line 354 (bus 85) greater than the possible maximum short-circuit current at the end (bus 86), as shown in \eqref{eq:iset>imax},
\begin{equation}
    I_{set}=KI^{Graph~D3QN}>I_{k,max}^{Global~Enum}
    \label{eq:iset>imax}
\end{equation}
where K is taken as 1.2, $I^{Graph~D3QN}$ is the maximum short-circuit current of line 354 calculated by Graph D3QN, and $I_{k,max}^{Global~Enum}$ is the actual maximum short-circuit current obtained by the brute-force enumeration. Otherwise, $I_{set}$ will fall into the range where a fault occurs on line 355, i.e., $P_1$ and $P_2$ will operate simultaneously in case of a fault on line 355, which don't satisfy the protection selectivity. The protection selectivity is tested on all operating conditions by stopping the other transmission lines in turn in the range N-2 for a total of 14,029 operating conditions. The results show that 100\% of the requirements of \eqref{eq:iset>imax} are satisfied in all operating conditions, i.e., the proposed Graph D3QN can fully satisfy the protection selectivity requirements. The above experiments are also carried out on 10 other transmission lines, and the results show that 100\% of the protection selectivity is satisfied.

\section{Conclusion}
\label{sec:CONCLUSION}
In this paper, a novel deep reinforcement learning-based method is proposed to enable a fast search for Extreme Operating Conditions (EOCs) in relay protection setting calculation. Named Graph Dueling Double Deep Q Network (Graph D3QN), the proposed framework employs graph neural networks to extract power system information and uses deep reinforcement learning for optimal policy making. A Guided Learning and Free Exploration (GLFE) training framework is also constructed to improve the convergence stability. The effectiveness and scalability of the proposed method are demonstrated on the IEEE 39-bus and 118-bus systems. Comparisons with existing EOC search methods show that the proposed Graph D3QN can achieve acceptable accuracy and can satisfy protection selectivity while significantly accelerating the computation speed. It is worth mentioning that the proposed method can be easily extended to solve other similar EOC search problems, only by strategically building appropriate environment and reward functions according to the problem requirements. Future research will consider improving the generalizability and scalability of the method and applying it to more applications that need to search for EOCs.

\bibliographystyle{IEEEtran}
\bibliography{IEEEabrv,main}

\end{document}